# Discovering Lexical Similarity Through Articulatory Feature-based Phonetic Edit Distance

**Tafseer Ahmed\*, Muhammad Suffian Nizami†, Muhammad Yaseen Khan\***

Mohammad Ali Jinnah University, Karachi\*, FAST NUCES, Faisalabad†
tafseer.ahmed@jinnah.edu, m.suffian@nu.edu.pk, yaseen.khan@jinnah.edu

**Abstract**

Lexical Similarity (LS) between two languages uncovers many interesting linguistic insights such as genetic relationship, mutual intelligibility, and the usage of one's vocabulary into other. There are various methods through which LS is evaluated. In the same regard, this paper presents a method of Phonetic Edit Distance (PED) that uses a soft comparison of letters using the articulatory features associated with them. The system converts the words into the corresponding International Phonetic Alphabet (IPA), followed by the conversion of IPA into its set of articulatory features. Later, the lists of the set of articulatory features are compared using the proposed method. As an example, PED gives edit distance of German word *vater* (/faːtər/) and Persian word 'پدر' (*pidar*, /pedær/) as 0.82; and similarly, Hebrew word 'שלום' (*shalom*, /ʃəlɒm/) and Arabic word 'سلام' (*salam*, /səlaːm/) as 0.93, whereas for a juxtapose comparison, their IPA based edit distances are 4 and 2 respectively. Experiments are performed with six languages (Arabic, Hindi, Marathi, Persian, Sanskrit, and Urdu). In this regard, we extracted part of speech wise word-lists from the Universal Dependency corpora and evaluated the LS for every pair of language. Thus, with the proposed approach, we find the genetic affinity, similarity, and borrowing/loan-words despite having script differences and sound variation phenomena among these languages.

**Keywords:** Lexical Similarity, Edit Distance, Phonetic Matching, Articulatory Features

## 1. Introduction

The present-day world invigorates many tasks in the field of computational linguistics for both: exploration and reevaluation. Many languages, which are once expected in isolation, are converted into global languages as people need them as their second language. This bilingualism is the phenomenon of the global village, colonial rule, fast communication technologies, etc. (Genesee, 2008; Errington, 2001; Burke, 2000). Thus, in the modern world, it is quite challenging to assume that a language retains vocabulary along with the heavy influence of other languages. For computation and evaluation, the Lexical Similarity (LS) can tell us how close are the two languages. There are various reasons for the languages being similar, for example, genetic affinity or borrowing of words. In the first case, the languages belonging to the same language family have common words inherited from an ancestor language. Such common words are called cognates.

The cognates are not identical to each other. With time, the form of the word changes due to the phenomenon of sound change. For example, consider the word for father in different Indo-Aryan languages, English: *father* /fɑ ðər/, German: *vater* /faːtər/, Persian: پدر (*pidar*, /pedær/), and Hindi: पिता (*pita*, /pɪ.t̪ɑː/). These words originating from the Proto–Indo–European (PIE) language's word pəter are not identical, as their form has changed due to the process of sound change occurring in hundreds of years. Hence, the simple method of string matching does not work correctly to find the cognates and lexical similarity.

In this paper, we present a string matching algorithm that is based on the articulatory features of the matched words. After devising the Phonetic Edit Distance (PED) method, we have shown the results by applying it to compare the lexical similarity of six different languages Arabic, Hindi, Marathi, Persian, Sanskrit, and Urdu. These languages engage different aspects of linguistic features such as writing scripts, Devanagari (for Sanskrit, Hindi, and Marathi) and Perso-Arabic (for Arabic, Persian, and Urdu); and mutual intangibility (such as Urdu and Hindi are mutually intelligible languages (Kelker, 1968; King, 2001)).

The significant contributions of this paper are one: proposing a string matching method on the basis of articulatory features; two: modifying Edit Distance (ED) method to deal with a soft matching of similar sounds, and three: using the proposed phonetic edit distance (PED) to find part of speech (PoS) wise lexical similarity of languages written in different scripts. The aforementioned third contribution is related to the work of Nizami et al. (2019). However, this paper presents an enhancement of the earlier work that was done on a mere three languages of Perso-Arabic script i.e., Arabic, Urdu, and Persian, barely computing edit distance on the orthographic transcription of words.

The organization of the paper is as follows: The literature review for both: string matching and phonetic (string) matching, and lexical similarities of the languages are discussed in section 2, the detailed discussion on the proposed PED is in section 3. Section 4 shares the computation of lexical similarities in the aforementioned languages, followed by the evaluation of results in section 5 and bibliographical references in the end.

## 2. Literature Review

This section respectively covers the literature review of the string and phonetic matching algorithms, as well as approaches for lexical similarity calculation.

### 2.1 String and Phonetic Matching

Research work contributed by Ukkonen (1985) stands as the premier and the most famous amongst all edit-based string matching algorithms. It gives the number of operations (insertion, deletion, or replacement) required to transform a string to another string. For example, the strings *fax* and *axe* have edit distance 2 as we perform two operations i.e., deleting 'f' of the first string and then inserting 'e' at its end for transforming it into the second string. There are variations in the ED algorithm to make it adept for the different tasks in Natural Language Processing (NLP) and Computational Biology (CB) e.g., spell correction (Ristad and Yianilos, 1998) string alignment (Jiang, 2002). However, its main shortcoming is that it takes to consider the letters used in the string as discrete-distinct units. In the comparison, either two letters are entirely similar (in this case no/zero operation is needed to transform 'p' into another 'p'), or they are entirely different (one operation is required to transform 'p' into 'b').

In contrast to the letter matching algorithms, Zobel and Dart (1996) presented a phonetic matching algorithm that gives the similarity of strings based on sounds of corresponding letters. Similarly, the Soundex algorithm has provided six equivalence classes of letters. The algorithm discards the vowels and transforms the consonants in the string into their mapped phonetic class (except the first letter), respectively (Yannakoudakis and Fawthrop, 1983). Following this scheme, both *robert* and *rupert* are transformed to the same string "r163" as both 'b' and 'p' belong to the same sound class {'b', 'f', 'p' and 'v'} having encoding value '1'. Philips (2000) introduces double Metaphone, which is widely employed for spell checking applications.

The works of Daitch–Mokotoff (DM Soundex) (Lait and Randell, 1996) and Beider-Morse Phonetic Matching (BMPM) (Beider, 2008) are the rule-based algorithms that also utilize the encoding scheme for matching the similar-sounding/homophonic names written in different languages. For example, the following are the same word spelled differently in different languages; *Schwarz* in German, *Szwarc* in Polish, *Şvarţ* in Romanian, *Chvarts* in French, and *Шварц* in Russian. Hence, these algorithms, through the ED, can be used to find the cognates according to the phonetic similarity. However, extending these algorithms to other languages is difficult as there are many languages and scripts. Thus, we specifically need an algorithm that applies phonetic matching without providing the equivalence classes of letters of similar sound toward defining acoustic-phonetic equivalence for vowels., as suggested in work by Broad (1976), of homophonic letters and complicated rules of spelling.

As mentioned above, the Soundex algorithm considers the letters. 'b', 'f', 'p' and 'v'. This similarity exists due to the similarity in the articulation of these sounds. The chart in figure 1 shows the IPAs arranged according to the place and manner of articulation. The sounds 'p' and 'b' are similar because they have the same place of articulation i.e. Bilabial (Scott and Ringel, 1971), and similarity of articulation i.e. Plosive (Roach, 1979). Further, the difference between these sounds are another articulatory feature i.e. *voicing* (Kuhl and Miller, 1975), for example, 'b' is a voiced consonant and 'p' is an unvoiced consonant. There are other features e.g. *aspirated* and *Pharyngeal* etc. for which the IPA are represented by diacritical marks. Hence, we cannot consider IPA symbols as atomic. An IPA symbol (or a sound) can be represented by a set of features.

|  | Bilabial | Labiodental | Dental | Alveolar | Postalveolar | Retroflex | Palatal | Velar | Uvular | Pharyngeal | Glottal |
|---|---|---|---|---|---|---|---|---|---|---|---|
| Plosive | p b |  |  | t d |  | ʈ ɖ | c ɟ | k g | q ɢ |  | ʔ |
| Nasal | m | ɱ |  | n |  | ɳ | ɲ | ŋ | ɴ |  |  |
| Trill | ʙ |  |  | r |  |  |  |  | ʀ |  |  |
| Tap or Flap |  | ⱴ |  | ɾ |  | ɽ |  |  |  |  |  |
| Fricative | ɸ β | f v | θ ð | s z | ʃ ʒ | ʂ ʐ | ç ʝ | x ɣ | χ ʁ | ħ ʕ | h ɦ |
| Lateral fricative |  |  |  | ɬ ɮ |  |  |  |  |  |  |  |
| Approximant |  | ʋ |  | ɹ |  | ɻ | j | ɰ |  |  |  |
| Lateral approximant |  |  |  | l |  | ɭ | ʎ | ʟ |  |  |  |

Symbols to the right in a cell are voiced, to the left are voiceless. Shaded areas denote articulations judged impossible.

Figure 1: Chart for the IPA, showing pulmonic consonants. Courtesy International Phonetic Association[1].

## 2.2 Lexical Similarities

Since the languages inherit words form a common ancestor language, therefore, it is quite evident that languages of the same language family have many homophonic words for the same concept. The difference in the sound of these words are widely studied and it is found that the change of the sound is systematic.

Grimm's Law deals with sound change in Germanic languages (for example English, German, Dutch, and Swedish, etc.) (Kortlandt,1988). It presented the observation that (some of them) voiced stops of the older language change to voiceless stops, and similarly, voiceless stops change into voiceless fricatives. There are other similar studies e.g. Dahl's Law (Lombardi, 1995) and Verner's Law (Page, 1998). Consider the example of the word *father* originating from /pəter/ in Proto–Indo–European (PIE). It is different among different Indo-European languages due to the change in sounds. It became /fader/ in Proto–Germanic, followed by changing into /faðer/ for the English language.

Nerbonne and Heeringa (1997) showed the dialect distances between two text by using the Levenshtein distance, cosine similarity, Hamming distance, and ASCII code based hashing. Kondrak (2001) presented a method to identify the cognates in the languages with inter-related vocabulary sets. It supports that the language similarity should be measured with phonetic multi-valued features, instead of orthographic measures like the longest common subsequence ratio.

Do et al. (2009) presented an approach namely, WNSim, for similarity analysis of words w.r.t their synsets in WordNet for a specific PoS. They also presented the lexical level matching (LLM) technique by combining word-level similarity to compute phrase level and sentence level similarity. Similarly, a string level similarity computation for identification of source code reported Kaur and Maini (2019), using the Rabin-Karp rolling hashing algorithm that outperforms various other algorithms. Gomaa and Fahmy (2013) performed a survey

---

[1] https://www.internationalphoneticassociation.org/

on three types of text similarity, i.e., *string-based*, *corpus-based* and *knowledge-based*.

Dijkstra et al. (2010) conducted an experiment about the cross-language similarity based on English and Dutch cognate sets, for example, English: *lamp* /læmp/, Dutch: *lamp* /lɑmp/ and English: *flood* /flʌd/, Dutch: *vloed* /vlut/. These cognate sets are useful for the bilingual translations. Similarly, Schepens (2013) showed cross-language distribution of cognates based on phonetics and high frequencies. The results show that cognate frequency reduced in less-closely related language-pairs as compared to more closely-related language-pairs. García and Souza (2014) conducted and experiment to compute the LS for the English and Purtugese language on 500 high frequency words. The genetic difference of both languages is less than 30%.

Likewise Dijkstra et al. (2010) and Schepens (2013), Carrasco-Ortiz (2019) reveals that the phonological and orthographic similarity matters in finding the cross-lingual cognates. It shows the calculation of the orthographic and phonetic similarity by using the Levenshtein distance, and found that both groups got benefit by orthographic similarity in reading Spanish and English words. Similarly, in another study, authors used six datasets for linking records across the languages like English and French, they evaluated their record linkages from DBpedia knowledge-base (Lehmann et al., 2015; Çakal et al., 2019)

Khan (2015) analyzed the historical background related to the origin of French and Urdu words. Although it showed the similarity of words on the basis of semantic, phonetic and etymology, and concluded the existence of genetic affinity between Urdu and French language, but it lacks computational model. Siew and Vitevitch (2019) did a similar work to proposed work in which the phonological and orthographic similarity structures of English words are characterized in a network of language. in which the links are made between the words orthographically and phonologically similar. Nizami et al. (2019) showed some work on the lexical similarity of three languages written in Arabic script. It employed the orthographic transcription technique for the word lists (set of lemmas) that share the same part of speech. The word lists mapped into IPA w.r.t the language, followed by computing their edit distance through Levenshtein distance.

## 3. "Phonetic" Edit Distance

We have an overview on phonetic similarity in §2, where the methods chiefly depend on the hand-crafted rules for specific or a set of languages. In contrast to those methods, in this section, one of the major contributions of this paper i.e., a method to find phonetic distance of two strings (already encoded in IPA) is presented and discussed in detail.

The Phonetic Edit Distance (PED) method takes two IPA strings as input and returns the phonetic distance between them. Likewise the standard ED, if the strings are same then the distance is zero, thus, in our case if the sounds are the same then the phonetic distance between them is zero. Otherwise, if there is a mismatch, then the resulted distance depends upon not only on the operations of insertion and deletion, but it also depends on the phonetic similarity of the sounds that are replaced.

In the standard ED, the distance of IPA strings /pɛn/ and /bɛnd/ is $\Phi(/pɛn/, /bɛnd/) = 2$, where $\Phi$ denotes function for standard ED, as the second string is formed by replacing 'p' with 'b' (bearing the operational cost 1) and inserting 'd' in the end (again, bearing the operational cost 1), hence the sum of all operational costs equals 2. However, with the proposed method, the PED for the same pair of IPA strings is $1 < \Psi < 2$, where $\Psi$ denotes the PED, as the operational cost of replacement is not fixed (as 1), and/since it varies between $(0, 1)$ depending on the phonetic similarity of the replaced sounds. Hence, as 'p' and 'b' in the given IPA strings, are more similar in articulation (as discussed in §2.1), their replacement cost would be less (for example, 0.2 w.r.t the proposed system) than the cost of replacing 'p' by 'z' (i.e., 0.35 w.r.t the result of the proposed system).

The major building blocks of this method are further discussed in §3.1–3.3.

### 3.1 Articulatory Features

The proposed system (or framework), as we have stated above, will work on the IPA; for which the words of the language will first be converted into IPA.

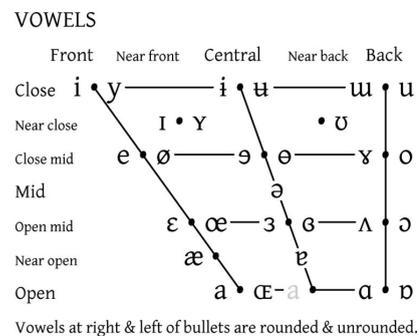

Figure 2: IPA chart (with articulation) of Vowels. Courtesy International Phonetic Association.

As we need to find the distance of sounds of the IPA string, we preferred the creation of features with continuous values. For *vowels*, we have two continuous features, *back* and *open*, (see the labels at the top, and left in figure 2) and one binary feature *rounded*. The *rounded* vowels are placed at right in the pairs of vowels in figure 2. Further, the feature *back* represents all labels (i.e., {*front, near-front, central, near-back,* and *back*}); and the values corresponding to it are set in a range $[0, 1]$, such that the *front* = 0, *near-front* = 0.25, *central* = 0.50, *near-back* = 0.75, and *back* = 1. Similarly, the feature *open* represents all labels (i.e., {*close, near-close, close-mid, mid, open-mid, near-open,* and *open*}); and, likewise the feature *back*, the values corresponding to *open* are set in a range $[0, 1]$, such that the *close* = 0, *near-close* = 0.17, *close-mid* = 0.33, *mid* = 0.50, *open-mid* = 0.67, *near-open* = 0.83, and *open* = 1. Lastly, the value for the feature *rounded*, which shows the binary characteristic, is either of the {1, 0} corresponding to the roundedness or non-roundedness of vowel. For example, consider the vowel 'i', and w.r.t the vowel chart presented in figure 2, we can set *open* = 0, *back* = 0 and *rounded* = 0. Similarly, the vowel 'u' has *open* = 0, *back* = 1 and *rounded* = 1. The vowel 'ɛ' has *back* = 0.25, *open* = 0.67 (as it is a *near-front* and *open-mid* vowel) and *rounded* = 1.

For *consonants*, we proposed the following features (and data-type of their respective values as) *place* (continuous), *manner* (discrète), *voiced* (binary), *airflow* (discrète), *aspirated* (binary) and *pharyngeal* (binary).

The feature *place* represents the place of articulation. As these places are present on the continuum inside human mouth, therefore, we assign relative positions as the value of these features. Hence the bilabial position (lips) have the value of 0.05, dental position (teeth) have the value of 0.15, and the glottal position (throat) has the value of 0.95. The other features have discrete or binary values.

There are two meta features, *label* and *type*, along with these articulatory features. The feature *label* has the IPA of the sound and the feature *type* encodes whether the sound is a consonant or a vowel.

### 3.2 Finding Distance of Sound

The articulatory features presented in §3.1 are crafted in such a way that the similar sounds have similar feature values. Also, we have different approaches for vowel–vowel and consonant–consonant comparison. The consonants are not compared with vowels and vice versa.

In order to compare two vowels: we employed Manhattan Distance (Craw, 2017), and assigned equal weight to all of the features under the same type, such as all non-binary features are weighted ⅔ of the distance, and binary features are weighted ⅓ of the distance. Consider algorithm 1 for the vowel comparison, where $w$ and $x$ are the features for the vowels, provided $w \neq x$, Manhattan Distance is denoted by $\Theta(\cdot)$; $\delta_{ob}$ and $\delta_r$ are the variable representing the distances of continuous and binary features.

---

**Algorithm 1** Phonetic Difference for Vowel

0:    PDV (w , x):
1:      $\delta_{ob} \leftarrow \Theta(\langle w_{open}, w_{back}\rangle, \langle x_{open}, x_{back}\rangle)$
2:      **if** $\delta_{ob} > 0.5$ **then**
3:        $\delta_r \leftarrow \Theta(\langle w_{rounded}\rangle, \langle x_{rounded}\rangle)$
4:        **return** $\frac{1}{3} \cdot (\delta_{ob} + \delta_r)$
5:      **else**
6:        **return** $\frac{1}{3} \cdot (\delta_{ob} + 1)$
7:      **endif**

---

Similarly for consonants comparison, as shown in algorithm 2, we gave ⅔ weight to the *place* and *manner* features. The remaining ⅓ weight is allocated to the other features. The feature *voiced* has ⅕ weight. The remaining features have the remaining weight represented as β which is $1 - \frac{2}{3} - \frac{1}{5}$. Currently in the proposed system, we have *airflow* and *aspirated*, as remaining features. However, the list can be extended without reducing the weight of the three main features. Moreover, *place* and *manner* features are more significant than any other feature having binary (or tertiary) values. Hence, we include the distance of other features only when the distance added by *place* and *manner* ($\delta_{m+p}$) is less than or equal to the threshold α (currently ½ of $\delta_{m+p}$). If the combined distance is more than this threshold α then we return that distance (scaling it as out of 1) without adding the distance of other features. We have a rule-based distances for the feature *manner*, hence, the lookup in the dictionary ξ is made, which results the distance when the key $\langle w_{manner}, x_{manner}\rangle$ is given. Lastly, $\sim *$, as mentioned in line 9, shows the Manhattan Distance for all of the remaining features.

---

**Algorithm 2** Phonetic Difference for Consonant

0:    PDC (w , x):
1:      $\delta_m \leftarrow \xi[\langle w_{manner}, x_{manner}\rangle]$
2:      $\delta_p \leftarrow \Theta(\langle w_{placed}\rangle, \langle x_{placed}\rangle)$
3:      $\delta_{m+p} \leftarrow \delta_m + \delta_p$
4:      **if** $\delta_{m+p} > \alpha$ **then**
5:        **return** $\delta_{m+p}$
6:      **else**
7:        $\delta_{m+p} \leftarrow \delta_{m+p} \cdot \frac{2}{3}$
8:        $\delta_v \leftarrow \Theta(\langle w_{voiced}\rangle, \langle x_{voiced}\rangle) \cdot \frac{1}{5}$
9:        $\delta_{rf} \leftarrow \Theta(\langle w_{\sim *}\rangle, \langle x_{\sim *}\rangle) \cdot \beta$
10:       **return** $\delta_{m+p} + \delta_v + \delta_{rf}$
11:     **endif**

---

### 3.3 Modification in Edit Distance

We made a modification in edit distance to account the phonetic similarity of sounds. The cost of insertion and deletion remains 1 (as in the standard ED). However, the replacement cost is calculated by calling the function PDV or PDC that is described in algorithm 1 and 2 respectively, while for algorithm 3 it is generalized as *phonetic_difference*($\cdot$). These functions return a real number between [0,1] i.e., 0 for the same sound and 1 for entirely different sounds. The pseudocode for the modified form of Phonetic Edit Distance (PED) is given below.

---

**Algorithm 3** Phonetic Edit Distance

0:    PED (source, target):
1:      s ← be the length of source IPA string
2:      t ← be the length of target IPA string
3:      **if** min( s , t ) = 0
4:        return max( s , t )
5:      **endif**
6:      **if** source[ s-1 ] = target[ t-1 ]
7:        cost = 0
8:      **else**
9:        ins_cost ← PED(source[ 0 : s-1 ] , target)+1
10:       del_cost ← PED(source , target[0 : t-1 ])+1
11:       rep_cost ← PED(source[ 0 : s-1 ], target[ 0 : t-1 ]
                   + phonetic_difference(source[ s-1 ] , target[ t-1 ])
12:     **endif**
13:     return min( ins_cost, del_cost, rep_cost )

Let us take two examples of calculating the PED for the words in different languages written in different scripts, Hebrew and Arabic are both Semetic languages, and hence many cognates. One of the cognate pairs is the greeting words 'שלום' (shalom, /ʃəlɒm/) and 'سلام' (salam, /səla:m/). The list of articulatory features corresponding to these words are given in table 1 and 2 respectively.

|  |  | IPA Symbols and Values | | | | |
|---|---|---|---|---|---|---|
| **Meta Features** | *label* | ʃ | ə | l | ɒ | m |
|  | *type* | *c* | *v* | *c* | *v* | *c* |
| **Phonetic/ Articulatory Features** | *method* | 0 | NA | 0 | NA | 0 |
|  | *place* | .45 | NA | .25 | NA | .05 |
|  | *manner* | fr | NA | ap | NA | ns |
|  | *voice* | 0 | NA | -1 | NA | -1 |
|  | *aspirated* | 0 | NA | 0 | NA | 0 |
|  | *open* | NA | 0.5 | NA | 1 | NA |
|  | *back* | NA | 0.5 | NA | 1 | NA |
|  | *rounded* | NA | 0 | NA | 1 | NA |

Table 1: Articulatory features corresponding to the Hebrew word 'שלום' (shalom, /ʃəlɒm/).

|  |  | IPA Symbols and Values | | | | |
|---|---|---|---|---|---|---|
| **Meta Features** | *label* | s | ə | l | a: | m |
|  | *type* | c | v | c | v | c |
| **Phonetic/ Articulatory Features** | *method* | 0 | NA | 0 | NA | 0 |
|  | *place* | .25 | NA | .25 | NA | .05 |
|  | *manner* | fr | NA | ap | NA | ns |
|  | *voice* | 1 | NA | -1 | NA | -1 |
|  | *aspirated* | 0 | NA | 0 | NA | 0 |
|  | *open* | NA | 0.5 | NA | 1 | NA |
|  | *back* | NA | 0.5 | NA | 0 | NA |
|  | *rounded* | NA | 0 | NA | 0 | NA |

Table 2: Articulatory features corresponding to the Arabic word 'سلام' (salam, /səla:m/).

In the IPA based strings, there are two mismatches, i.e., the consonant 'ʃ' is to be replaced by the consonant 's'. In the standard ED (Φ), it will be cost of one operation, but with the application of PED on the feature lists of these consonant, the resulting PED (Ψ) was 0.267. Similarly, the Ψ between the vowels 'ɒ' and 'a:' is 0.667. Hence, the Ψ (of articulatory features) of these two strings is 0.267 + 0.667 = 0.934 (despite having two replacements).

Similarly, the PED of German word *vater* (/faːtər/) and Persian word 'پدر' (*pidar*, /pedær/) having four replacements, i.e., 'f'→'p' = 0.1, 'a:'→'e' = 0.223, 't'→'d' = 0.217, and 'ə'→'æ' = 0.277, is calculated as 0.817.

## 4. Experimental Setup

In this section, firstly we discuss the requirements and their reasons for which this experiment is designed. Then we show the systematic approach of calculating PoS-wise lexical similarity between the two languages.

LS between two languages is calculated on the basis of count/ratio of similar words present in two languages. There are three major issues in calculation of the lexical similarity. *one*—How do we decide whether the two compared words are similar or not? *two*—How many words will be compared? *three*—Do we compare any of the words or do we choose the words on the basis of some criterion?

The proposed PED method gives reply to the first question. The method does not require to manually decide whether two words are cognate are similar or not. The PED gives edit distance as a measure of similarity, and word lists can be aligned using this measure. A method of alignment and lexical similarity calculation is presented in §4.2.

The second and third questions are about choosing the appropriate word lists. Nizami et al. (2019) proposed the calculation of LS on the basis of different parts of speech (PoS). As annotated UD corpora are available for more than 60 languages, and all of those corpora uses the same pos tagset. Thus, it is a good choice to run an experiment with UD corpora. As mentioned earlier, languages have similar words due to various reasons, some word classes e.g. pronouns and numbers are similar due to genetic affinity, hence we can say that languages belonging to the same language family or sub-family have cognates due to genetic affinity (i.e., inheritance from the parent or ancestor language). The other word classes e.g. nouns and proper nouns may have a significant influence due to borrowing from a genetically unrelated or distant language. Hence, PoS-wise similarity will portray different aspects of the LS and this gives the reply to the second question. As we choose to extract PoS-wise word lists from the UD coropra, count of words to be compared, the count depends on the words present in the corpora. As manually selecting the words from long lists (e.g. of nouns and verbs) is not feasible, we use all the words extracted from the lists.

### 4.1 Languages, Scripts, Corpora, and PoS Tags

For this experiment, we chose six languages that are written in two entirely different scripts, i.e., Perso-Arabic and Devanagari. The reasons behind choosing these languages are: *first*—the Universal Dependency (UD) corpora in considerable size of these six languages are available; *secondly*—the conversion system of text-to-IPA for these languages/scripts is created. The six languages involved in the experiments are Arabic, Persian, and Urdu (all written in Arabic script (Kachru, 1990; Rangila et al., 2001)), as well as Hindi, Marathi, and Sanskrit (all written in Devanagari script (Kachru, 1990)). The PoS tags involved in this experiment are: adposition, auxiliary, coordinating and subordinating conjunctions, determiner, particles, pronouns, nouns, proper nouns, and verbs.

## 4.2 Calculating Lexical Similarity

The important components of our experiment are systematically described in the following sub-sections.

### 4.2.1 Creating Word Lists

The UD corpora are available on the website[2] of Universal Dependencies. The corpora has the tagged information of dependency structures modeled in CoNLL-U format (Çöltekin et al., 2017). We extracted PoS-wise word lists by processing these structures. The dependency structures have *lemma* corresponding to the word e.g. the lemma *like* corresponds to the words *like*, *likes*, *liked*, and *liking*. Thus, we chose lemma instead of word, as it reduces the size of word list and the need of comparison of words in different lists. Hence, the word-lists and words mentioned in the following text are actually lemma-lists and lemmas respectively. However, we retain the use the term 'word' in the following discussion.

### 4.2.2 Word-to-IPA Conversion

The next component of the system converts the word into corresponding IPA string. It is considerable that the word-to-IPA conversion is available for many languages, as most of the paper/online dictionaries have IPA entry corresponding to the word. Moreover, many languages have a simple one to one letter to IPA mapping. Hence we infer that IPA strings are easier to obtain and entertain in this system.

We implemented a small module for the *orthographic* word-to-IPA conversion for Arabic and Devanagari script. Since the Arabic script does not have short vowels, therefore, we omitted these too in the Devanagari script conversion, as it is to be compared with Arabic script. This module is an ad-hoc arrangement because we can have multiple methods for mapping of word to IPA in different languages, and even for the same language. Currently, we implement a mapper for script-to-IPA. However, since many languages have word-lists (and dictionaries) having IPA corresponding to the given word, we can use these word-IPA lists in the future extensions of this system.

### 4.2.3 IPA-to-Articulatory Features

The list of articulatory features for phonetic matching is already presented (in table 1 and 2) in §3.3. A one-to-one mapping is required to convert the IPA string into a list of list of articulatory features. Currently, we have articulatory features for IPA used in these 6 languages. However, the list can be extended easily by listing features corresponding to the remaining IPA symbols.

### 4.2.4 Computing Lexical Similarities

The components described above give us (PoS-wise) list of words of different languages. Now, for the lexical similarity, consider the two word-lists, $L_1$ and $L_2$; $L_1 \neq L_2$, entertaining the same PoS. Further, we arrange the $L_1$ and $L_2$ such that $|L_1| < |L_2|$. Algorithm 4 shows the comparison of words in $L_1$ and $L_2$ to find the LS in two languages. The words 'w' and 'x', as shown in step 3 and 4 of algorithm 4, encompass articulatory features. Further, if we do not normalize the PED resulting value (as shown in step 5 of algorithm 4) by the maximum length between word of $L_1$ and $L_2$, then the comparison of smaller words get less PED value in comparison to the words having larger length.

---

**Algorithm 4** Lexical Similarity between two lists.

```
0:   L₁ ← be the list of language 1.
1:   L₂ ← be the list of language 2.
2:   Ψ_all ← 0
3:   for each w in L₁:
4:       for each x in L₂:
5:           Ψ[x] ← PED( w, x ) · 1/max(|w|, |x|)
6:       Ψ_all ← Ψ_all + min_a(Ψ[a])
7:       Remove a from L2
8:   μΨ ← Ψ_all ÷ |L₁|
```
---

The minimum value of edit distance as result of comparison of a word w of L1 with all the words x of L2 one by one is added to the $\Psi_{all}$ that have the overall value of edit distance. The word with minimum value in this step is removed from the list L2, so it is not used again in the comparison. Dividing the Ψ by the length of $L_1$ (as shown in step 8) gives the average per letter PED (μΨ) of the two lists, if this μΨ = 0 then the lists are identical, similarly, if it is 1 then the lists are entirely different. A smaller value (closer to 0) show that most of the words are the same or similar in sound. The larger values (closer to 1) show that most of the words in the two lists are different.

## 5. Results and Discussion

We compared PoS wise word lists of six languages using the algorithms described above, and the results are presented in figure 3. These languages have genetic as well as social affinities with each other. Arabic belongs to the Semetic family of languages. Persian belongs to Iranian branch of the Indo–Europen→Indo–Iranian languages. Urdu, Hindi, Marathi, and Sanskrit belong to Indo–Aryan branch of the Indo–Europen→Indo–Iranian languages. Sanskrit is an ancient language; however, the other three are modern languages. Urdu has much social interaction with Arabic and Persian, so it has borrowed vocabulary and phonetics from these languages; otherwise, Hindi and Urdu are different variants of the same language. Arabic, Persian, and Urdu are written in Arabic script, while Hindi, Marathi, and Sanskrit are written in the Devanagari script.

Figure 3 depicts the PoS wise lexical similarity using heatmaps. The lighter color shows a lower value of PED per letter and hence, higher similarity. The darker color shows larger value of PED per letter and hence, lower similarity. We did not calculate the similarity of some pairs when there are less than 5 words in one of the lists. The tiny or empty lists are not suitable for inferring some results.

---

[2] https://universaldependencies.org/

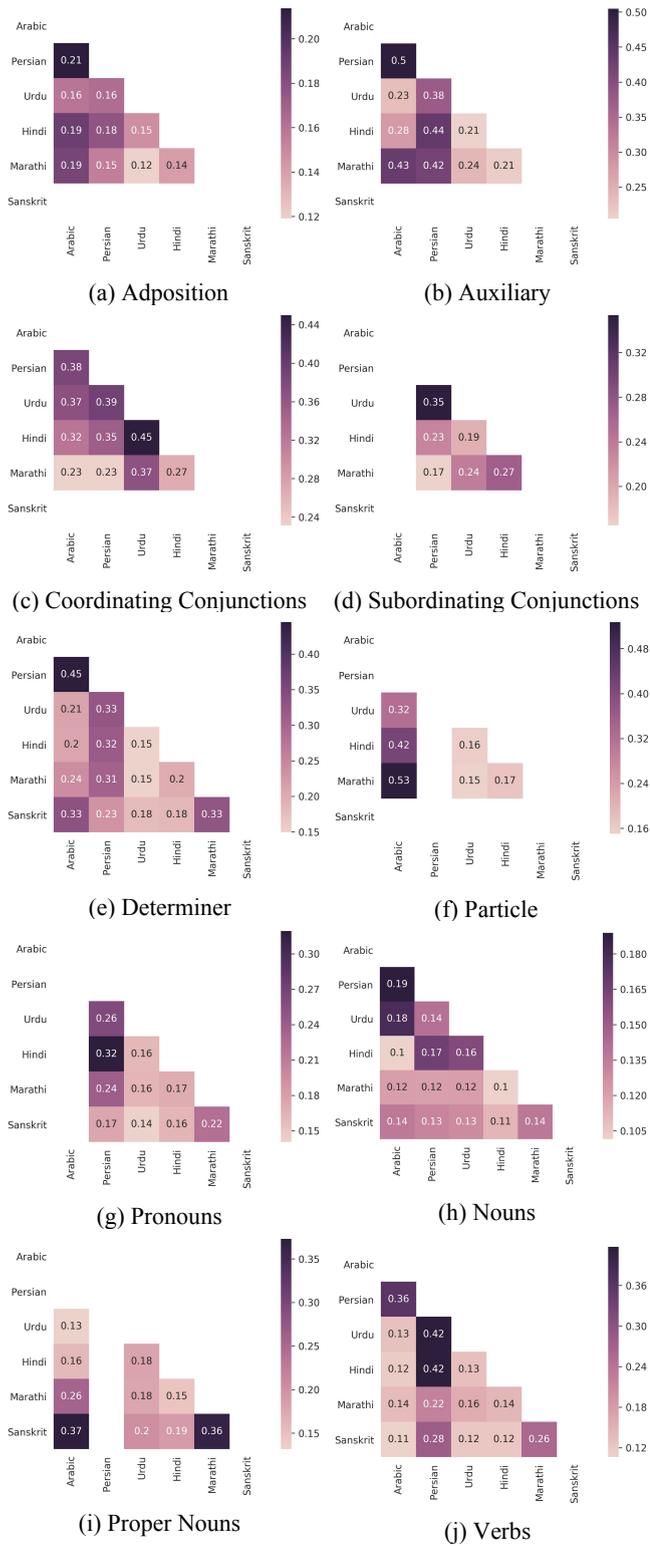

(a) Adposition  (b) Auxiliary
(c) Coordinating Conjunctions  (d) Subordinating Conjunctions
(e) Determiner  (f) Particle
(g) Pronouns  (h) Nouns
(i) Proper Nouns  (j) Verbs

Figure 3: Part of speech wise lexical similarity of languages, i.e., Arabic, Persian, Urdu, Hindi, Marathi, and Sankskrit.

We find that for five out of ten PoS, the top two similar languages (i.e., having lowest PED) are Indo Aryan languages. These languages w.r.t to PoS are: for adpositions Urdu–Marathi and Hindi–Marathi, for auxiliaries Urdu–Hindi and Hindi–Marathi, for determiners Urdu–Marathi and Urdu–Hindi, for particles Urdu–Marathi and Urdu–Hindi, and for pronouns Urdu–Sanskrit and Urdu–Hindi. All of these PoS are the closed-class, i.e. new words are usually not added in these lists.

For the remaining PoS, we have Persian–Marathi and Arabic–Marathi for coordinating conjunctions; Persian–Marathi and Urdu–Hindi for Subordinating conjunctions. These, too, belong to the closed-class of words, and these anomalous results need explanation. Persian and Marathi belong to the same sub-family of Indo-Iranian languages; however we expect closer affinity of Marathi with Hindi or Urdu.

Further, we find Arabic–Hindi and Hindi–Marathi for nouns, and Urdu–Arabic and Hindi–Marathi, as the most similar (i.e., having lowes PED) languages for proper nouns. These PoS are open-class, and we expect borrowing from Arabic for the words of these PoS.

Amonst all, the verb shows the most irrelevant result. We find substantial similarity of Arabic verbs with the Indo-Aryan languages. We suspect that the smaller length of root in the Arabic verb is the reason for this anomaly. Arabic lanugage has three-letter roots that is a very potential candidate word for an easy match with any of the root/lemma of the compared language.

Although the results of the proposed methodology has a limitation of *false positives*. Thus, we assume that the average of the false positives will be the same in all of the pairs. This assumption holds for the majority of the PoS wise language pairs. However, it may need some modification for two small words or too small word lists.

When we browsed through the content of different PoS wise word lists, we found that the parallelism of similar design principles does not hold in some cases. The count of pronouns in Urdu, Hindi, and Marathi are 110, 63, and 24, respectively. Similarly, the count of Urdu, Hindi, and Marathi auxiliaries are 78, 49, and 19, respectively. There should not be such a difference in closed-class word lists of closely related languages. The reason is excluding a subclass of words as they give PoS or putting the inflected form as the lemma in the CoNLL-U structures.

## 6. Conclusion and Future Work

We presented an algorithm for articulatory feature-based phonetic edit distance (PED). This algorithm helps to identify the cognates that have different spellings, and IPA mapping is different among languages. We used the PED to find the lexical similarity of PoS wise lists of different pairs of languages. Most of the calculated similarities are in agreement with the genetic affinity of the compared languages. Hence, the method can be used on a more extensive set of languages after removing the following limitations.

The current work used a mapping of the letter(s) to IPA. A better approach is the usage of digitally available lexicographic resources, e.g. dictionaries, etc. It will resolve the many matters of ambiguity, e.g. letter(s) to IPA, silent letters, and unwritten diacritic marks.

One can work on a better set of corpora or better cleaning, as we found some problems of (non-)parallel design, implementation errors, and cleaning with UD corpora. We may some manually or automatically tagged parallel corpus, e.g. Europarl parallel corpus or Wikipedia dumps. However, this method has the problem of inaccurate tagging by the PoS tagger.

## 7. Bibliographical References